\documentclass[sigconf]{acmart}

\usepackage{bm}
\usepackage[noend]{algpseudocode}
\usepackage{algorithmicx,algorithm}
\usepackage{graphicx}

\AtBeginDocument{%
  \providecommand\BibTeX{{%
    \normalfont B\kern-0.5em{\scshape i\kern-0.25em b}\kern-0.8em\TeX}}}

\copyrightyear{2021}
\acmYear{2021}
\begin{document}

\fancyhead{}

\title{Capsule Graph Neural Networks with EM Routing}

\author{Yu Lei and Jing Zhang}
\email{jzhang@njust.edu.cn}
\authornote{Dr. Jing Zhang is the corresponding author.}
\affiliation{%
  \institution{School of Computer Science and Engineering, Nanjing University of Science and Technology}
  \streetaddress{200 Xiaolingwei Street}
  \city{Nanjing}
  \state{Jiangsu}
  \country{China}
  \postcode{210094}
}

\renewcommand{\shortauthors}{Lei and Zhang}

\begin{abstract}
To effectively classify graph instances, graph neural networks need to have the capability to capture the part-whole relationship existing in a graph. A capsule is a group of neurons representing complicated properties of entities, which has shown its advantages in traditional convolutional neural networks. This paper proposed novel Capsule Graph Neural Networks that use the EM routing mechanism (CapsGNNEM) to generate high-quality graph embeddings. Experimental results on a number of real-world graph datasets demonstrate that the proposed CapsGNNEM outperforms nine state-of-the-art models in graph classification tasks.
\end{abstract}




%%  \includegraphics[width=\textwidth]{sampleteaser}
%%  \caption{Seattle Mariners at Spring Training, 2010.}
%%  \Description{Enjoying the baseball game from the third-base
%%  seats. Ichiro Suzuki preparing to bat.}

\maketitle

\section{Introduction}
Recent years have witnessed the increasing attention to Graph Neural Networks (GNNs) \cite{Scarselli2009TheGN,Bruna2014SpectralNA,Kipf2017SemiSupervisedCW,Velickovic2018GraphAN}, which have demonstrated remarkable advantages in many tasks performing on graph-structured data, such as node classification \cite{Kipf2017SemiSupervisedCW,Rong2020DropEdgeTD,Gao2019SemiSupervisedGE,Zhou2021MultilabelGN}, graph classification \cite{Bacciu2018ContextualGM,Li2019SemiSupervisedGC}, and link prediction \cite{Zhang2018LinkPB,Ragunathan2020LinkPB}. GNNs are generalized from traditional deep-learning models like CNNs and RNNs (handling grid or linear-structured data such as images and sequences), which can take graph-structured instances as their input. Many real-world data like social networks exhibit complicated graph structures. GNNs can be directly applied to the original graph structures, learning more effective representation.

This paper focuses on graph convolutional neural networks (GCNs), which employ convolutional operations for more general graph-structured data. The construction of GCNs can be categorized into spectral approaches and spatial approaches \cite{Liu2020Intro}. The common principle of both spectral and spatial convolutions is to recursively update node embeddings by aggregating information from topological neighbors, which can capture the local structure of nodes. The learned embeddings of nodes can be used in various tasks such as node classification and link prediction. To classify a whole graph instance, we need further to learn a graph embedding (a higher-level embedding compared with the node embedding), which is achieved by pooling methods \cite{Zhang2018AnED,Gao2019Unet,Bianchi2020SpectralCW,Yuan2020StructPool}. However, most existing graph pooling methods have two weaknesses: 1) The pooling operation may map different graphs or nodes into the same embedding, resulting in the model being incapable to capture meaningful information. 2) The pooling methods only consider the topology of the graph but ignore the part-whole relationships. For example, a molecule has a hydroxyl on its left and right sides, respectively. If they are not linked with each other, they will be clustered into different groups. However, because they have the same chemical properties they should be in the same group.

To solve the above issues, a group of neurons in the networks, namely \textit{capsule}, was introduced, which encodes the activate probability of an entity as well as reserves the detailed properties of the entity such as position, direction, connection, and so on. Capsule neural networks have shown their effectiveness in modeling part-whole relationship on images \cite{Hinton2011TransformingA,Sabour2017DynamicR,Hinton2018MatrixCW}.
It solves the viewpoint-invariant problem by transformation matrices. Capsule networks employ a routing mechanism to generate high-level capsules by the voting of low-level capsules. Compared with pooling methods that only reserve activated features, a routing mechanism preserves all information from low-level capsules and routes it to the closest high-level capsules. The most popular routing mechanism is \textit{dynamic routing by agreement} \cite{Sabour2017DynamicR}, which uses the length of a capsule to represent its degree of salience. Capsule and dynamic routing have been introduced in GNNs. CapsGNN \cite{Zhang2018Capsule} stacks the node features extracted by GCNs to build capsules and uses dynamic routing and attention mechanisms to generate high-level graph capsules as well as class capsules \cite{Zhang2018Capsule}. GCAPS-CNN~\cite{Verma2018GC} uses capsules to capture the highly informative output in a small vector in place of a scaler output currently employed in GCNs. The capsule is consists of higher-order statistical moments, which is permutationally invariant and can be computed via fast matrix multiplication. Another routing mechanism representing a capsule as matrices is \textit{matrix capsule with EM routing} \cite{Hinton2018MatrixCW}. Using matrices instead of vectors to represent capsules can reduce the size of conversion matrices between capsules. For example, converting a $3\times3$ matrix to a $3\times6$ matrix requires a $3\times6$ conversion matrix, while in dynamic routing converting a 9-dimensional vector to an 18-dimensional matrix requires a $9\times18$ conversion matrix. Besides, the probability of an entity in EM routing is represented by a parameter $a$ (instead of a vector in dynamic routing), which avoids using the squashing function to ensure that the length of the vector is in a feasible range.

To the best of our knowledge, the matrix capsules with EM routing have never been introduced into GNNs. In this paper, we propose novel Capsule Graph Neural Networks with EM routing (CapsGNNEM), which uses node features extracted from GCN to generate high-quality graph embeddings by EM routing. Experimental results on a number of real-world graph datasets demonstrate the advantages of the proposed CapsGNNEM in graph classification.

\section{Preliminaries} \label{sec:Preliminary}
We first formulate graph classification and then briefly introduce graph neural networks and capsule neural networks.

\paragraph{Graph Classification} A graph is represented by $G=(V, X, A)$, where $V=\{v_1, v_2, v_3, \cdots, v_N\}$ is a set of $N$ nodes, $X\in \mathbb{R}^{N\times C}$ is the feature matrix for nodes with feature channels $C$, and $A\in \{0, 1\}^{N\times N}$ is the adjacency matrix. If there is an edge from $v_i$ to $v_j$, then we have $A_{ij}=1$, otherwise, $A_{ij}=0$. Given a set of labeled graphs $D=\{(G_0, y_0), (G_1, y_1), (G_2, y_2), \cdots\}$, we aim to learn a graph embedding $g(G)\in \mathbb{R}^k$ for each graph, which encodes its attributes and structural information. Then, the graph representation (embeddings) can be used for graph classification.

\paragraph{Graph Neural Networks} Graph neural networks usually follow a neighborhood aggregation fashion to learn a node representation by applying a neighbor aggregation function to the representations of its neighbor nodes after propagation. More specifically, a neighbor aggregation function for the $i$-th node has the form:
\begin{equation}
h_i^{(l+1)}=f(h_i^{(l)},\{h_v^{(l)}|v\in\mathcal{N}(v)\}),
\end{equation}
where $\mathcal{N}(v)$ is a set of neighbors of node $i$ and $h_i^{(l)}$ is the representation of node $i$ at layer $l$. In each layer, the representation of every node is updated by the neighbor aggregation function. In graph classification, we need the representation of a whole graph, which usually can be obtained by simply summing up or averaging all the node embeddings on the graph. It obviously cannot capture the structural information of graphs. In capsule neural networks, graph embeddings are calculated by a hierarchical stack of capsules. The lowest-level capsules are the matrix presentation of nodes.

\paragraph{Capsule Neural Networks} A \textit{capsule} is a group of neurons whose outputs represent different properties of an entity such as pose, deformation, texture, etc \cite{Hinton2011TransformingA}. Routing algorithms are critical to the formation of different levels of capsules. In \textit{dynamic routing by agreement}, the activation  $\in([0, 1)$ of a capsule is represented by its revised length after being calculated by a \textit{squashing} function. The \textit{squashing} function ensures that a short vector shrinks to a length slightly greater than $0$ and a long vector shrinks to a length slightly less than $1$, defined as  $v_j=\frac{\left\lVert s_j\right\rVert^2 }{1+\left\lVert s_j\right\rVert^2 }\frac{s_j}{\left\lVert s_j\right\rVert }$, where $s_j$ and $v_j$ are the capsules before and after squashing, respectively. Accordingly, \textit{EM routing} defines a special value as the activation value of a capsule, representing the certainty of an entity \cite{Hinton2018MatrixCW}, which is of better interpretability than using the capsule length as its activation value. (The details of EM routing are in Section~\ref{Subsec:CCL}). Since the higher-level capsules are obtained through the voting mechanism by the lower-level capsules, the hierarchical capsules can model the relationship between the part and whole.

\section{The Proposed Method} \label{Sec:Method}
This section presents the proposed CapsGNNEM in details.

\subsection{Network Architecture} \label{Subsec:Arch}
\begin{figure}
	\centering
	\includegraphics[width=3.4in]{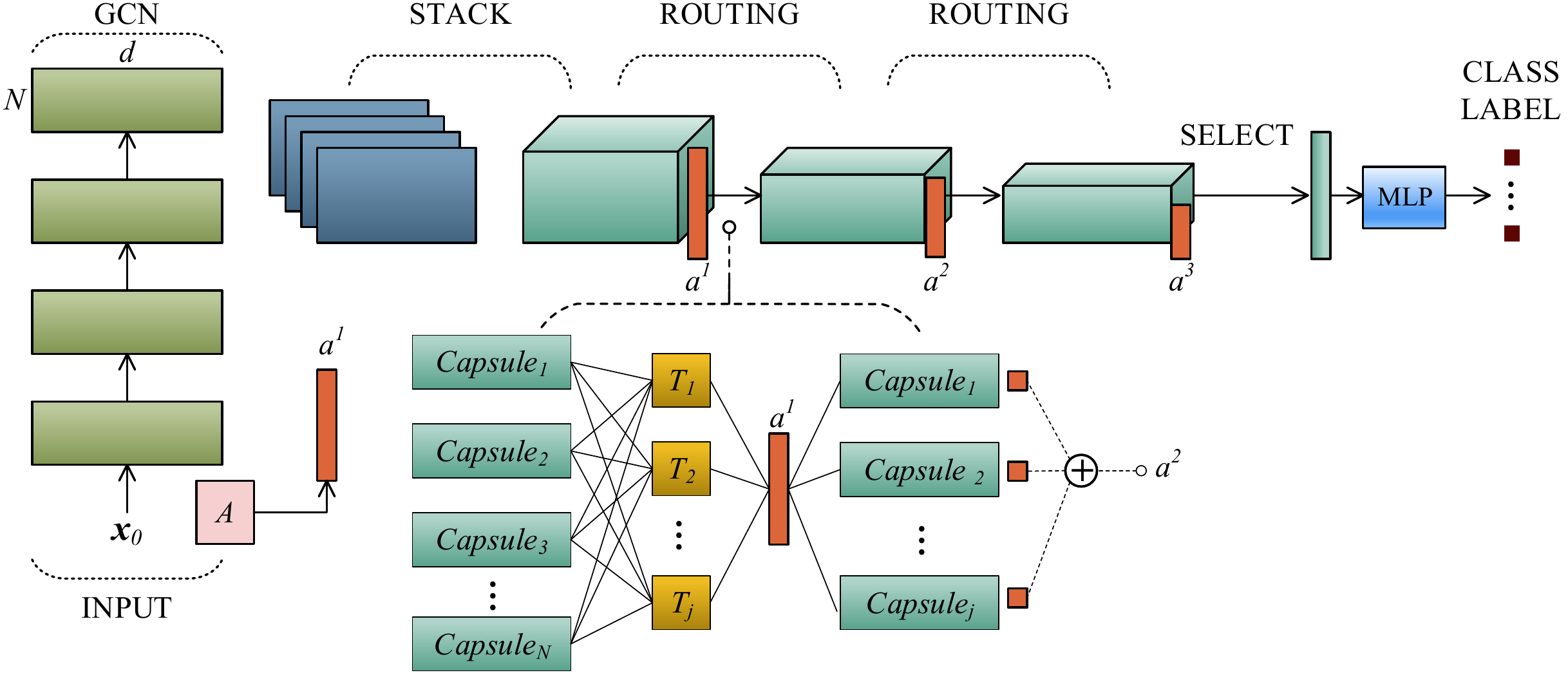}
	\caption{Architecture of the Proposed GapsGNNEM \label{fig:Arch}}
	\vspace{-0.45cm}
\end{figure}

Figure \ref{fig:Arch} shows the architecture of our GapsGNNEM. It consists of three key components: \textit{Primary Capsule Layer}, \textit{Capsule Convolutional Layer}, and \textit{Readout Layer}. In the first component, it uses GCN to extract different receptive-field node features to form the primary capsules and uses adjacency to form the initial activations. In the second component, the EM routing is applied to obtain high-level graph capsules. Finally, in the third component, the capsule with the best activation representing the most significant embeddings is chosen to make predictions by a multilayer perceptron (MLP).

\subsection{Primary Capsule Layer} \label{Subsec:PrimCaps}
The goal of the primary capsule layer is to generate a capsule for each node on a graph. The features of a node are initialized with its original features. For those non-attribute graphs, we can choose Local Degree Profile \cite{Cai2018ASY} as the node attributes. The same as articles \cite{Zhang2018Capsule,Verma2018GC}, our method uses the GCN model (the commonest GNNs) \cite{Kipf2017SemiSupervisedCW} to aggregate information from a node's local neighborhood at a lower layer. The aggregation function at layer $l+1$ is as follows:
\begin{equation}
X^{(l+1)}=f(\tilde{D}^{-1/2}\tilde{A}\tilde{D}^{-1/2}X^{(l)}\Theta),
\end{equation}
where $f$ is a nonlinear activation function, $\tilde{A}=A+I_{N}$ and $\tilde{D}_{ij}=\sum_j\tilde{A}_{ij}$, $X^{(l)}\in \mathbb{R}^{N\times d}$ represents the $d$-dimensional features for all $N$ nodes at layer ${l}$ (particularly, $X^{(0)}=X$), and $\Theta\in \mathbb{R} ^{d\times d^{\prime}}$ is a matrix of filter parameters that shrinks the dimension of node features from $d$ to $d^\prime$. As mentioned in the introduction, the matrix representation for a node is more efficient than vector representation. Thus, we stack the node features from different GCN layers to generate a matrix representation of features, which is consists of different receptive-field information. For example, if we have $k (\ge 3)$ GCN layers, the initial capsule for node $i$ is a $k\times d^{\prime}$ matrix $[X^{(1)}_{i:}; X^{(2)}_{i:};\cdots; X^{(k)}_{i:}]$. The EM routing also requires an activation value. The bigger the degree of a node, the more central and important it is. Therefore, for node $i$, we intuitively choose its degree as the initial activation value, i.e., $a_i=\frac{D_i}{\sum_{j} D_j}$, where $D_i$ is its degree.

\subsection{Capsule Convolutional Layer} \label{Subsec:CCL}
After obtaining primary capsules and their initial activation values, the capsule convolutional layer decides how to assign lower-level active capsules to higher-level capsules. This process can be viewed as clustering (solved by an EM algorithm). Each higher-level capsule corresponds to a cluster center and each lower-level active capsule corresponds to a data point (or a fraction of a data point if the capsule is partially activated). First, it captures the part-whole relationship for each lower-level capsule $c_i^{(l)}$ and higher-level capsule $c_j^{(l+1)}$. This relationship is measured by a transformation matrix, which is called \textit{voting} in EM routing and defined as follows:
\begin{equation}
V_{ij}^{(l)}=c_i^{(l)} T_{ij}^{(l)},
\end{equation}
where $V_{ij}^{(l)}\in \mathbb{R}^{k\times d^{\prime}}$ represents the voting result from capsule $i$ at layer $l$ for some capsule $j$ at layer $l+1$, 
$c_i^{(l)}$ represents capsule $i$ at layer $l$, and $T_{ij}^{(l)}$ is a transformation matrix. We use a simplified Gaussian Mixture Model (GMM) to generate higher-level capsules. That is, a higher-level capsule can be viewed as a center of multiple lower-level capsules. The simplified Gaussian distribution $\mathcal{N}(\bm{x};\bm{\mu},\Sigma)$ has a diagonal covariance matrix $diag(\bm{\sigma}^2)$. In this settings, the posterior probability of a voting $V_{ij}$ belonging to the $j$-th Gaussian (i.e., capsule $j$ at the higher level) is
\begin{equation}
R_{ij}=\frac{a_j\mathcal{N} (V_{ij};\bm{\mu}_j,diag(\bm{\sigma}_j^2))}{\sum_ja_j\mathcal{N} (V_{ij};\bm{\mu}_j,diag(\bm{\sigma}_j^2))},
\end{equation}
where activation $a_j$ for capsule $j$ is a mixture coefficient of GMM and $V_{ij}$ is treated as a $k*d^{\prime}$-dimentional vector. Because lower-level capsules vote for a higher-level capsule $j$ the contribution coeffient $r_{ij}$ of capsule $i$ when calculating cluster center (capsule) $j$ should consider its activation value $a_i$ as follows:
\begin{equation}
r_{ij} =\frac{a_iR_{ij}}{\sum_ia_iR_{ij}}.
\end{equation}
Then, the parameters of the $j$-th Gaussian (parameter $\bm{\mu}_j$ is capsule $j$) can be updated in an EM procedure as follows:
\begin{equation}
\bm{\mu}_j=\sum\limits_ir_{ij}V_{ij},
\end{equation}
\begin{equation}
(\sigma_j^h)^2=\sum\limits_ir_{ij}(V_{ij}^h-\mu_j^h)^2,
\end{equation}
where the superscript $h$ represents the $h$-th component of a vector. Following article \cite{Hinton2018MatrixCW}, the activation value for each capsule $j$ is calculated as follows:
\begin{align}
	cost_j^h &= \sum\limits_i-r_{ij}\ln(p_{i|j}^h)=\sum\limits_ir_{ij}(\ln(\sigma_j^h)+1/2+\ln(2\pi)/2),\nonumber\\
	a_j&=logistic\left(\lambda\left(\beta_a-\beta_u\sum\limits_ir_{ij}-\sum\limits_hcost_j^h\right)\right),
\end{align}
where $p_{i|j}$ is the probability of $V_{ij}$ given capsule $j$ (i.e., $p_{i|j}=\mathcal{N}(V_{ij};\bm{\mu}_j,diag(\bm{\sigma}_j^2))$), $cost_j$ represents the entropy of capsule $j$, $r_{ij}$ is the amount of votes assigned to capsule $j$, $\beta_a$ and $\beta_u$ are the learned parameters, and $\lambda$ is a hyper-parameter used to adjust the range of values to be better for the logistic function. Thus, $a_j$ represents the activation for capsule $j$ in the higher layer.

Finally, we summarize the whole EM routing in Algorithm~\ref{Algo:EMR}.
\begin{algorithm}
	\caption{EM routing algorithm returns activation values and capsules in layer $l+1$ given activation values and capsules in layer $l$. $\beta_u$ and $\beta_a$ are learned parameters and parameter $\lambda$ is fixed.\label{Algo:EMR}}
	\begin{algorithmic}[1]
		\Procedure{EM routing}{$a^{in}, X$}
		\State$R\gets 1/|\Omega_{l+1}|, V\gets\text{VOTING}(X)$
		\Comment{$\Omega_l$ stands for a set of capsules at layer $l$}
		\While{$t \le maxiter$}
		\State$a^{out}, \bm{\mu}, \bm{\sigma}\gets\text{M-STEP}(a^{in}, R, V)$
		\State$R\gets\text{E-STEP}(\bm{\mu}, \bm{\sigma}, a^{out}, V)$
		\EndWhile
		\State \textbf{Return} $a^{out}, \bm{\mu}$
		\EndProcedure
	\end{algorithmic}
	\begin{algorithmic}[1]
		\Function{VOTING}{$X$}
		\State$\forall i\in \Omega_{l},\forall j\in\Omega_{l+1}:$ calculate $V_{ij}$ by Eq. (3), where $c_i=X_i$
		\State \textbf{Return} $V$
		\EndFunction
	\end{algorithmic}
	\begin{algorithmic}[1]
		\Function{M-STEP}{$a^{in}, R, V, j$} 
		\State$\forall i\in \Omega_{l}:$ calculate $r_{ij}$ by Eq. (5)
		\State$\forall j\in \Omega_{l+1}:$ calculate $\bm{\mu}$, $\bm{\sigma}_j$, and $a_j^{out}$ by Eqs. (6), (7), and (8)
		\State \textbf{Return} $a^{out}$, $\bm{\mu}$, and $\bm{\sigma}_j$
		\EndFunction
	\end{algorithmic}
	\begin{algorithmic}[1]
		\Function{E-STEP}{$a^{out}, V, \bm{\mu}, \bm{\sigma}$}
		\State$\forall j\in \Omega_{l+1}:$ calculate $R_{ij}$ by Eq. (4), where $a_j=a^{out}$
		\State \textbf{Return} $R$
		\EndFunction
	\end{algorithmic}
\end{algorithm}

\subsection{Readout Layer}
The CNN-based capsule networks with EM routing \cite{Hinton2018MatrixCW} directly use the activation value of the last layer to optimize with the \textit{spread loss} function, which is very sensitive to hyper-parameter settings. Instead, we choose the capsule on the last layer with the best activation, which represents the most significant embedding. Then, we calculate the probability of each class using a fully connected layer and optimized it by the cross-entropy loss function. This will bring some robustness to our model when adjusting hyper-parameters.

\begin{table*}[htp]
	\caption{Comparison against nine state-of-the-art models on seven datasets in terms of accuracy (in percent)}
	\noindent\centering
	\begin{tabular}{lcccccccc}	
		$\mathbf{Algorithm}$ & $\mathbf{MUTAG}$ & $\mathbf{NCI1}$ & $\mathbf{PROTEINS}$ & $\mathbf{D\&D}$ & $\mathbf{COLLAB}$ & $\mathbf{IMDB-B}$ & $\mathbf{IMDB-M}$\\
		\hline 
		WL \cite{Shervashidze2011WeisfeilerLehmanGK}& $82.50 \pm 0.36$ & $82.19 \pm 0.18$ & $74.68 \pm 0.49$ & $79.78 \pm 0.36$ & $79.02 \pm 1.77$ & $73.40 \pm 4.63$ & $49.33 \pm 4.75$\\
		GK \cite{Shervashidze2009EfficientGK} & $81.58 \pm 2.11$ & $62.49 \pm 0.27$ & $71.67 \pm 0.55$ & $78.45 \pm 0.26$ & $72.84 \pm 0.28$ & $65.87 \pm 0.98$ & $43.89 \pm 0.38$\\      
		DGK \cite{Yanardag2015DeepGK} & $87.44 \pm 2.72$ & $80.31 \pm 0.46$ & $75.68 \pm 0.54$ & $73.50 \pm 1.01$ & $73.09 \pm 0.25$ & $66.96 \pm 0.56$ & $44.55 \pm 0.52$\\
		AWE \cite{Ivanov2018AnonymousWE} & $87.87 \pm 9.76$ & $78.18 \pm 3.02$ & $73.28 \pm 2.64$ & $71.51 \pm 4.02$ & $73.93 \pm 1.94$ & $74.45 \pm 5.83$ & $50.45 \pm 3.61$\\
		\hline
		GCN \cite{Kipf2017SemiSupervisedCW} & $87.20 \pm 5.11$ & $ 74.95 \pm 0.64$ & $75.65 \pm 3.24$ & $79.12 \pm 3.07$ & $\mathbf{81.72 \pm 1.64}$ & $73.30 \pm 5.29$& $49.37 \pm 2.21$\\
		PSCN \cite{Niepert2016LearningCN}  & $88.95 \pm 4.37$ & $76.34 \pm 1.68$ & $75.00 \pm 2.51$ & $76.27 \pm 2.64$ & $72.60 \pm 2.15$ & $71.00 \pm 2.29$ & $45.23 \pm 2.84$\\
		DGCNN \cite{Zhang2018AnED} & $85.83 \pm 1.66$ & $74.44 \pm 0.47$ & $75.54 \pm 0.94$ & $79.37 \pm 0.94$ & $73.76 \pm 0.49$ & $70.03 \pm 0.86$ & $47.83 \pm 0.85$\\
		\hline
		GCAPS-CNN  \cite{Verma2018GC} & $84.12 \pm 3.19$ & $\mathbf{82.72 \pm 2.38}$ & $76.40 \pm 4.17$ & $77.62 \pm 4.99$ & $77.71 \pm 2.51$ & $71.69 \pm 3.40$ & $48.50 \pm 4.10$\\
		CapsGNN \cite{Zhang2018Capsule} & $86.67 \pm 6.88$ & $78.35 \pm 1.55$ & $76.28 \pm 3.63$ & $75.38 \pm 4.17$ & $79.62 \pm 0.91$ & $73.10 \pm 4.83$ & $50.27 \pm 2.65$\\
		\textbf{Ours (CapsGNNEM)} & $\mathbf{90.51 \pm 2.33}$ & $75.03 \pm 0.98$ & $\mathbf{77.41 \pm 1.12}$ & $\mathbf{81.51 \pm 4.31}$ & $75.23 \pm 1.47$ & $\mathbf{75.40 \pm 3.25}$ & $\mathbf{50.87 \pm 3.25}$\\
		\hline
	\end{tabular} \label{Table1}
\end{table*}

\section{Experiments}\label{sec:Exp}
In this section, we evaluate the performance of CapsGNNEM against a number of state-of-the-art graph classification methods.
\subsection{Methods in Comparison}
We compared the proposed CapsGNNEM with three categories of existing methods.

The first category is called \textit{kernel-based} methods, including
the Weisfeiler-Lehman subtree kernel (WL) \cite{Shervashidze2011WeisfeilerLehmanGK}, the graphlet count kernel (GK) \cite{Shervashidze2009EfficientGK}, the deep graph kernel (DGK) \cite{Yanardag2015DeepGK}, and anonymous walk embeddings (AWE) \cite{Ivanov2018AnonymousWE}, which learn graph representation based on the substructs defined by the kernel methods.

The second category is called \textit{GNN-based} methods. This category includes several methods. We selected five state-of-the-art GNN-based methods used in the experiments. GCN \cite{Kipf2017SemiSupervisedCW} is one of the most influential models on graph neural networks. PATCHY-SAN (PSCN) \cite{Niepert2016LearningCN} selects a sequence of nodes and generates node representation by a receptive field, then uses CNN for graph classification. Deep Graph CNN (DGCNN) \cite{Zhang2018AnED} first uses a sorted pooling method to replace global pooling based on the node feature extracted by GNN, then uses CNN and MLP for classification.

The last category is called \textit{capsule-based} methods. We included two capsule-based methods GCAPS-CNN \cite{Verma2018GC} and CapsGNN \cite{Zhang2018Capsule} in comparison. GCAPS-CNN uses higher-order statistical moments that are permutationally invariant to generate a capsule for each node, then computes covariance to deal with graph classification. CapsGNN generates capsules from multi-scale node embeddings extracted by GCN, then uses dynamic routing to generate graph embeddings for graph classification.

\subsection{Experimental Settings}
Seven real-world datasets were used in the experiments, which were derived from important application areas of graph classification, including four biological graph datasets (MUTAG, NCI1, PROTEINS, and D\&D ) and three social network datasets (COLLAB, IMDB-B, and IMDB-M). The same hyper-parameter settings of CapsGNNEM were used for all datasets. For the \textit{primary capsule layer}, we used 4 GCN layers to extract node features. The dimension of each node feature was set to 16. Thus, the dimension of a capsule after stacking is $4 \times 16$. For the \textit{capsule convolutional layer}, the number of output capsules is 3 times that of the following layer. The number of output capsules in the last layer is equal to the number of target classes. The number of iterations in the EM routing is set to 2 and the $\lambda$ is set to 0.1. We applied 10-fold cross-validation to evaluate the performance. For each evaluation process, the original instances were randomly partitioned into a training set and a test set. Then, we performed 10-fold cross-validation on the training set. In each cross-validation process, one fold of the training set was used to adjust and evaluate the performance of the trained model and the remainder was used to train a model. We selected the model with the best validation accuracy as a representative. The instances in the test set were predicted against the representative and the performance in terms of accuracy was evaluated. The above process was repeated 10 times, the average results and their standard deviations were reported.

\subsection{Experimental Results}
Table \ref{Table1} lists the comparison results on the seven datasets in terms of accuracy. For each dataset, the average accuracy of the best model is in bold. Overall, our proposed CapGNNEM outperforms the other state-of-the-art methods on the datasets MUTAG, PROTEINS, D\&D, IMDB-B, and IMDB-M. CapGNNEM wins on five out of seven datasets, which suggests that CapsGNNEM has the best overall performance. On three biological datasets (MUTAG, PROTEINS, and D\&D), compared with two capsule-based methods (GCAPS-CNN and CapsGNN), CapsGNNEM improves the classification accuracy by margins of $3.84\%$, $1.01\%$ and $3.89\%$, respectively. The reason is that CapsGNNEM represents node features in form of capsules that can capture information of graphs from different aspects instead of only one embedding used in other GNN-based approaches. This is helpful to retain the properties of a graph when generating graph capsules. Besides, the capsule can capture the same properties and ignore the difference in their substruct, which is more important in biological datasets. These results also consistent with the property of EM routing, as it focuses more on extracting properties from child capsules by voting and routing. However, the social network datasets do not have node attributes. Therefore, applying routing to all child capsules leads to a loss of structural information in the graph. Therefore, even if CapsGNNEM has a higher average performance than other methods on IMDB-B and IMDB-M, comparing with the second-best methods, its margins of exceeding ($0.95$ on IMDB-B and $0.42$ on IMDB-M) are still smaller than those obtained from biological datasets. Nonetheless, CapsGNNEM still demonstrates its strong capability to capture graph properties and part-whole relationships.

\section{Conclusion and Future Work}
This paper proposes a novel CapsGNNEM model for graph classification, which combines capsule structures and graph convolutional networks. The capsules with EM routing not only can capture the properties and part-whole relations of graphs from GCN-extracted node features but also reduce the complexity of networks. Experimental results on a number of datasets have confirmed the advantages of the proposed model.

At present, the capsule convolution layer only performs global convolutions, which does not make good use of the structural information of the graph. In the future, we will construct a kind of local convolutions, where a cluster of capsules with a strong structural relationship is computed together. Moreover, since the model can represent hierarchical structures of capsules, the readout layer can extract hierarchical features as its input.

\begin{acks}
This work has been supported by the National Key Research and Development Program of China under grant 2018AAA0102002 and the National Natural Science Foundation of China under grants 62076130 and 91846104.
\end{acks}

\clearpage

\bibliographystyle{ACM-Reference-Format}
\bibliography{sample-base}


\begin{thebibliography}{27}

%%% the Web address.

\ifx \showCODEN    \undefined \def \showCODEN     #1{\unskip}     \fi
\ifx \showDOI      \undefined \def \showDOI       #1{#1}\fi
\ifx \showISBNx    \undefined \def \showISBNx     #1{\unskip}     \fi
\ifx \showISBNxiii \undefined \def \showISBNxiii  #1{\unskip}     \fi
\ifx \showISSN     \undefined \def \showISSN      #1{\unskip}     \fi
\ifx \showLCCN     \undefined \def \showLCCN      #1{\unskip}     \fi
\ifx \shownote     \undefined \def \shownote      #1{#1}          \fi
\ifx \showarticletitle \undefined \def \showarticletitle #1{#1}   \fi
\ifx \showURL      \undefined \def \showURL       {\relax}        \fi
% invisible to TeX
\providecommand\bibfield[2]{#2}
\providecommand\bibinfo[2]{#2}
\providecommand\natexlab[1]{#1}
\providecommand\showeprint[2][]{arXiv:#2}

\bibitem[\protect\citeauthoryear{Bacciu, Errica, and Micheli}{Bacciu
  et~al\mbox{.}}{2018}]%
        {Bacciu2018ContextualGM}
\bibfield{author}{\bibinfo{person}{Davide Bacciu}, \bibinfo{person}{Federico
  Errica}, {and} \bibinfo{person}{Alessio Micheli}.}
  \bibinfo{year}{2018}\natexlab{}.
\newblock \showarticletitle{Contextual graph Markov model: A deep and
  generative approach to graph processing}. In \bibinfo{booktitle}{\emph{The
  35th International Conference on Machine Learning (ICML)}}.
  \bibinfo{pages}{294--303}.
\newblock


\bibitem[\protect\citeauthoryear{Bianchi, Grattarola, and Alippi}{Bianchi
  et~al\mbox{.}}{2020}]%
        {Bianchi2020SpectralCW}
\bibfield{author}{\bibinfo{person}{Filippo~Maria Bianchi},
  \bibinfo{person}{Daniele Grattarola}, {and} \bibinfo{person}{Cesare Alippi}.}
  \bibinfo{year}{2020}\natexlab{}.
\newblock \showarticletitle{Spectral clustering with graph neural networks for
  graph pooling}. In \bibinfo{booktitle}{\emph{The 37th International
  Conference on Machine Learning (ICML)}}. \bibinfo{pages}{874--883}.
\newblock


\bibitem[\protect\citeauthoryear{Bruna, Zaremba, Szlam, and LeCun}{Bruna
  et~al\mbox{.}}{2014}]%
        {Bruna2014SpectralNA}
\bibfield{author}{\bibinfo{person}{Joan Bruna}, \bibinfo{person}{Wojciech
  Zaremba}, \bibinfo{person}{Arthur Szlam}, {and} \bibinfo{person}{Yann
  LeCun}.} \bibinfo{year}{2014}\natexlab{}.
\newblock \showarticletitle{Spectral networks and locally connected networks on
  graphs}. In \bibinfo{booktitle}{\emph{The 2nd International Conference on
  Learning Representations (ICLR)}}.
\newblock


\bibitem[\protect\citeauthoryear{Chen and Wang}{Chen and Wang}{2018}]%
        {Cai2018ASY}
\bibfield{author}{\bibinfo{person}{Cai Chen} {and} \bibinfo{person}{Yusu
  Wang}.} \bibinfo{year}{2018}\natexlab{}.
\newblock \showarticletitle{A simple yet effective baseline for non-attribute
  graph classification}.
\newblock \bibinfo{journal}{\emph{ArXiv}}  \bibinfo{volume}{abs/1811.03508}
  (\bibinfo{year}{2018}).
\newblock


\bibitem[\protect\citeauthoryear{Gao and Ji}{Gao and Ji}{2019}]%
        {Gao2019Unet}
\bibfield{author}{\bibinfo{person}{Hongyang Gao} {and}
  \bibinfo{person}{Shuiwang Ji}.} \bibinfo{year}{2019}\natexlab{}.
\newblock \showarticletitle{Graph u-nets}. In \bibinfo{booktitle}{\emph{The
  36th International Conference on Machine Learning (ICML)}}.
  \bibinfo{pages}{2083--2092}.
\newblock


\bibitem[\protect\citeauthoryear{Gao, Zhang, and Zhou}{Gao
  et~al\mbox{.}}{2019}]%
        {Gao2019SemiSupervisedGE}
\bibfield{author}{\bibinfo{person}{Kaisheng Gao}, \bibinfo{person}{Jing Zhang},
  {and} \bibinfo{person}{Cangqi Zhou}.} \bibinfo{year}{2019}\natexlab{}.
\newblock \showarticletitle{Semi-supervised graph embedding for multi-label
  graph node classification}. In \bibinfo{booktitle}{\emph{International
  Conference on Web Information Systems Engineering}}.
  \bibinfo{pages}{555--567}.
\newblock


\bibitem[\protect\citeauthoryear{Hinton, Krizhevsky, and Wang}{Hinton
  et~al\mbox{.}}{2011}]%
        {Hinton2011TransformingA}
\bibfield{author}{\bibinfo{person}{Geoffrey~E. Hinton}, \bibinfo{person}{Alex
  Krizhevsky}, {and} \bibinfo{person}{Sida~D. Wang}.}
  \bibinfo{year}{2011}\natexlab{}.
\newblock \showarticletitle{Transforming auto-encoders}. In
  \bibinfo{booktitle}{\emph{International Conference on Artificial Neural
  Networks (ICANN)}}. \bibinfo{pages}{44--51}.
\newblock


\bibitem[\protect\citeauthoryear{Hinton, Sabour, and Frosst}{Hinton
  et~al\mbox{.}}{2018}]%
        {Hinton2018MatrixCW}
\bibfield{author}{\bibinfo{person}{Geoffrey~E. Hinton}, \bibinfo{person}{Sara
  Sabour}, {and} \bibinfo{person}{Nicholas Frosst}.}
  \bibinfo{year}{2018}\natexlab{}.
\newblock \showarticletitle{Matrix capsules with EM routing}. In
  \bibinfo{booktitle}{\emph{The 6th International Conference on Learning
  Representations (ICLR)}}.
\newblock


\bibitem[\protect\citeauthoryear{Ivanov and Burnaev}{Ivanov and
  Burnaev}{2018}]%
        {Ivanov2018AnonymousWE}
\bibfield{author}{\bibinfo{person}{Sergey Ivanov} {and} \bibinfo{person}{Evgeny
  Burnaev}.} \bibinfo{year}{2018}\natexlab{}.
\newblock \showarticletitle{Anonymous walk embeddings}. In
  \bibinfo{booktitle}{\emph{International Conference on Machine Learning,}}.
  \bibinfo{pages}{2186--2195}.
\newblock


\bibitem[\protect\citeauthoryear{Kipf and Welling}{Kipf and Welling}{2017}]%
        {Kipf2017SemiSupervisedCW}
\bibfield{author}{\bibinfo{person}{Thomas~N. Kipf} {and} \bibinfo{person}{Max
  Welling}.} \bibinfo{year}{2017}\natexlab{}.
\newblock \showarticletitle{Semi-supervised classification with graph
  convolutional networks}. In \bibinfo{booktitle}{\emph{The 5th International
  Conference on Learning Representations (ICLR)}}.
\newblock


\bibitem[\protect\citeauthoryear{Li, Rong, Cheng, Meng, Huang, and Huang}{Li
  et~al\mbox{.}}{2019}]%
        {Li2019SemiSupervisedGC}
\bibfield{author}{\bibinfo{person}{Jia Li}, \bibinfo{person}{Yu Rong},
  \bibinfo{person}{Hong Cheng}, \bibinfo{person}{Helen Meng},
  \bibinfo{person}{Wenbing Huang}, {and} \bibinfo{person}{Junzhou Huang}.}
  \bibinfo{year}{2019}\natexlab{}.
\newblock \showarticletitle{Semi-supervised graph classification: A
  hierarchical graph perspective}. In \bibinfo{booktitle}{\emph{The World Wide
  Web Conference (WWW)}}. \bibinfo{pages}{972--982}.
\newblock


\bibitem[\protect\citeauthoryear{Liu and Zhou}{Liu and Zhou}{2020}]%
        {Liu2020Intro}
\bibfield{author}{\bibinfo{person}{Zhiyuan Liu} {and} \bibinfo{person}{Jie
  Zhou}.} \bibinfo{year}{2020}\natexlab{}.
\newblock \showarticletitle{Introduction to graph neural networks}.
\newblock \bibinfo{journal}{\emph{Synthesis Lectures on Artificial Intelligence
  and Machine Learning}} \bibinfo{volume}{14}, \bibinfo{number}{2}
  (\bibinfo{year}{2020}), \bibinfo{pages}{1--127}.
\newblock


\bibitem[\protect\citeauthoryear{Niepert, Ahmed, and Kutzkov}{Niepert
  et~al\mbox{.}}{2016}]%
        {Niepert2016LearningCN}
\bibfield{author}{\bibinfo{person}{Mathias Niepert}, \bibinfo{person}{Mohamed
  Ahmed}, {and} \bibinfo{person}{Konstantin Kutzkov}.}
  \bibinfo{year}{2016}\natexlab{}.
\newblock \showarticletitle{Learning convolutional neural networks for graphs}.
  In \bibinfo{booktitle}{\emph{The 33rd International Conference on Machine
  Learning (ICML)}}. \bibinfo{pages}{2014--2023}.
\newblock


\bibitem[\protect\citeauthoryear{Ragunathan, Selvarajah, and Kobti}{Ragunathan
  et~al\mbox{.}}{2020}]%
        {Ragunathan2020LinkPB}
\bibfield{author}{\bibinfo{person}{Kumaran Ragunathan},
  \bibinfo{person}{Kalyani Selvarajah}, {and} \bibinfo{person}{Ziad Kobti}.}
  \bibinfo{year}{2020}\natexlab{}.
\newblock \showarticletitle{Link prediction by analyzing common neighbors based
  subgraphs using convolutional neural network}. In
  \bibinfo{booktitle}{\emph{The 24th European Conference on Artificial
  Intelligence (ECAI)}}. \bibinfo{pages}{1906--1913}.
\newblock


\bibitem[\protect\citeauthoryear{Rong, Huang, Xu, and Huang}{Rong
  et~al\mbox{.}}{2020}]%
        {Rong2020DropEdgeTD}
\bibfield{author}{\bibinfo{person}{Yu Rong}, \bibinfo{person}{Wenbing Huang},
  \bibinfo{person}{Tingyang Xu}, {and} \bibinfo{person}{Junzhou Huang}.}
  \bibinfo{year}{2020}\natexlab{}.
\newblock \showarticletitle{DropEdge: Towards deep graph convolutional networks
  on node classification}. In \bibinfo{booktitle}{\emph{The 8th International
  Conference on Learning Representations (ICLR)}}.
\newblock


\bibitem[\protect\citeauthoryear{Sabour, Frosst, and Hinton}{Sabour
  et~al\mbox{.}}{2017}]%
        {Sabour2017DynamicR}
\bibfield{author}{\bibinfo{person}{Sara Sabour}, \bibinfo{person}{Nicholas
  Frosst}, {and} \bibinfo{person}{Geoffrey~E. Hinton}.}
  \bibinfo{year}{2017}\natexlab{}.
\newblock \showarticletitle{Dynamic routing between capsules}. In
  \bibinfo{booktitle}{\emph{NIPS}}.
\newblock


\bibitem[\protect\citeauthoryear{Scarselli, Gori, Tsoi, Hagenbuchner, and
  Monfardini}{Scarselli et~al\mbox{.}}{2009}]%
        {Scarselli2009TheGN}
\bibfield{author}{\bibinfo{person}{Franco Scarselli}, \bibinfo{person}{Marco
  Gori}, \bibinfo{person}{Ah~Chung Tsoi}, \bibinfo{person}{Markus
  Hagenbuchner}, {and} \bibinfo{person}{Gabriele Monfardini}.}
  \bibinfo{year}{2009}\natexlab{}.
\newblock \showarticletitle{The graph neural network model}.
\newblock \bibinfo{journal}{\emph{IEEE Transactions on Neural Networks}}
  \bibinfo{volume}{20} (\bibinfo{year}{2009}), \bibinfo{pages}{61--80}.
\newblock


\bibitem[\protect\citeauthoryear{Shervashidze, Schweitzer, Leeuwen, Mehlhorn,
  and Borgwardt}{Shervashidze et~al\mbox{.}}{2011}]%
        {Shervashidze2011WeisfeilerLehmanGK}
\bibfield{author}{\bibinfo{person}{Nino Shervashidze}, \bibinfo{person}{Pascal
  Schweitzer}, \bibinfo{person}{Erik Jan~Van Leeuwen}, \bibinfo{person}{Kurt
  Mehlhorn}, {and} \bibinfo{person}{Karsten Borgwardt}.}
  \bibinfo{year}{2011}\natexlab{}.
\newblock \showarticletitle{Weisfeiler-Lehman graph kernels}.
\newblock \bibinfo{journal}{\emph{Journal of Machine Learning Research}}
  \bibinfo{volume}{12} (\bibinfo{year}{2011}), \bibinfo{pages}{2539--2561}.
\newblock


\bibitem[\protect\citeauthoryear{Shervashidze, Vishwanathan, Petri, Mehlhorn,
  and Borgwardt}{Shervashidze et~al\mbox{.}}{2009}]%
        {Shervashidze2009EfficientGK}
\bibfield{author}{\bibinfo{person}{Nino Shervashidze},
  \bibinfo{person}{S.~V.~N. Vishwanathan}, \bibinfo{person}{Tobias Petri},
  \bibinfo{person}{Kurt Mehlhorn}, {and} \bibinfo{person}{Karsten Borgwardt}.}
  \bibinfo{year}{2009}\natexlab{}.
\newblock \showarticletitle{Efficient graphlet kernels for large graph
  comparison}. In \bibinfo{booktitle}{\emph{Artificial Intelligence and
  Statistics}}. \bibinfo{pages}{488--495}.
\newblock


\bibitem[\protect\citeauthoryear{Veličković, Cucurull, Casanova, Romero,
  Liò, and Bengio}{Veličković et~al\mbox{.}}{2018}]%
        {Velickovic2018GraphAN}
\bibfield{author}{\bibinfo{person}{Petar Veličković},
  \bibinfo{person}{Guillem Cucurull}, \bibinfo{person}{Arantxa Casanova},
  \bibinfo{person}{Adriana Romero}, \bibinfo{person}{Pietro Liò}, {and}
  \bibinfo{person}{Yoshua Bengio}.} \bibinfo{year}{2018}\natexlab{}.
\newblock \showarticletitle{Graph attention networks}. In
  \bibinfo{booktitle}{\emph{The 6th International Conference on Learning
  Representations (ICLR)}}.
\newblock


\bibitem[\protect\citeauthoryear{Verma and Zhang}{Verma and Zhang}{2018}]%
        {Verma2018GC}
\bibfield{author}{\bibinfo{person}{Saurabh Verma} {and} \bibinfo{person}{Zhi-Li
  Zhang}.} \bibinfo{year}{2018}\natexlab{}.
\newblock \showarticletitle{Graph capsule convolutional neural networks}. In
  \bibinfo{booktitle}{\emph{Joint ICML and IJCAI Workshop on Computational
  Biology}}.
\newblock
\urldef\tempurl%
\url{arXiv:1805.08090}
\showURL{%
\tempurl}


\bibitem[\protect\citeauthoryear{Yanardag and Vishwanathan}{Yanardag and
  Vishwanathan}{2015}]%
        {Yanardag2015DeepGK}
\bibfield{author}{\bibinfo{person}{Pinar Yanardag} {and}
  \bibinfo{person}{S.~V.~N. Vishwanathan}.} \bibinfo{year}{2015}\natexlab{}.
\newblock \showarticletitle{Deep graph kernels}. In
  \bibinfo{booktitle}{\emph{Proceedings of the 21th ACM SIGKDD International
  Conference on Knowledge Discovery and Data Mining}}.
  \bibinfo{pages}{1365--1374}.
\newblock


\bibitem[\protect\citeauthoryear{Yuan and Ji}{Yuan and Ji}{2020}]%
        {Yuan2020StructPool}
\bibfield{author}{\bibinfo{person}{Hao Yuan} {and} \bibinfo{person}{Shuiwang
  Ji}.} \bibinfo{year}{2020}\natexlab{}.
\newblock \showarticletitle{StructPool: Structured graph pooling via
  conditional random fields}. In \bibinfo{booktitle}{\emph{The 8th
  International Conference on Learning Representations (ICLR)}}.
\newblock


\bibitem[\protect\citeauthoryear{Zhang and Chen}{Zhang and Chen}{2018a}]%
        {Zhang2018LinkPB}
\bibfield{author}{\bibinfo{person}{Muhan Zhang} {and} \bibinfo{person}{Yixin
  Chen}.} \bibinfo{year}{2018}\natexlab{a}.
\newblock \showarticletitle{Link prediction based on graph neural networks}. In
  \bibinfo{booktitle}{\emph{The 32nd Annual Conference on Neural Information
  Processing Systems (NeurIPS)}}.
\newblock


\bibitem[\protect\citeauthoryear{Zhang, Cui, Neumann, and Chen}{Zhang
  et~al\mbox{.}}{2018}]%
        {Zhang2018AnED}
\bibfield{author}{\bibinfo{person}{Muhan Zhang}, \bibinfo{person}{Zhicheng
  Cui}, \bibinfo{person}{Marion Neumann}, {and} \bibinfo{person}{Yixin Chen}.}
  \bibinfo{year}{2018}\natexlab{}.
\newblock \showarticletitle{An end-to-end deep learning architecture for graph
  classification}. In \bibinfo{booktitle}{\emph{The 32nd AAAI Conference on
  Artificial Intelligence (AAAI)}}. \bibinfo{pages}{4438--4445}.
\newblock


\bibitem[\protect\citeauthoryear{Zhang and Chen}{Zhang and Chen}{2018b}]%
        {Zhang2018Capsule}
\bibfield{author}{\bibinfo{person}{Xinyi Zhang} {and} \bibinfo{person}{Lihui
  Chen}.} \bibinfo{year}{2018}\natexlab{b}.
\newblock \showarticletitle{Capsule graph neural network}. In
  \bibinfo{booktitle}{\emph{The 6th International Conference on Learning
  Representations (ICLR)}}.
\newblock


\bibitem[\protect\citeauthoryear{Zhou, Chen, Zhang, Li, Hu, and Sheng}{Zhou
  et~al\mbox{.}}{2021}]%
        {Zhou2021MultilabelGN}
\bibfield{author}{\bibinfo{person}{Cangqi Zhou}, \bibinfo{person}{Hui Chen},
  \bibinfo{person}{Jing Zhang}, \bibinfo{person}{Qianmu Li},
  \bibinfo{person}{Dianming Hu}, {and} \bibinfo{person}{Victor Sheng}.}
  \bibinfo{year}{2021}\natexlab{}.
\newblock \showarticletitle{Multi-label graph node classification with label
  attentive neighborhood convolution}.
\newblock \bibinfo{journal}{\emph{Expert Systems with Applications}}
  \bibinfo{volume}{180} (\bibinfo{year}{2021}), \bibinfo{pages}{115063}.
\newblock


\end{thebibliography}

%%\appendix

\end{document}